\documentclass[final]{cvpr}
\pdfoutput=1
\usepackage{times}
\usepackage{epsfig}
\usepackage{graphicx}
\usepackage{amsmath}
\usepackage{amssymb}
\usepackage{subcaption}
\usepackage{booktabs}
\usepackage{array}
\usepackage{xspace}
\usepackage{cases}
\usepackage{multirow}
\usepackage{url}
\usepackage{xcolor}
\usepackage{paralist}
\usepackage[normalem]{ulem}
\usepackage{enumitem}

\newcommand{\mycaption}[2]{\caption{\textbf{#1.}~#2}}
\usepackage{pifont}
\newcommand{\cmark}{\ding{52}}%
\newcommand{\xmark}{\ding{56}}%

\usepackage[pagebackref=True,breaklinks=true,colorlinks,bookmarks=false]{hyperref}
\graphicspath{ {Fig/} }

\def\cvprPaperID{10} 
\def\confYear{CVPR 2021}
\begin{document}

\title{Video-based Person Re-identification without Bells and Whistles}
\vspace{-2mm}
\author{
Chih-Ting Liu\textsuperscript{1}, Jun-Cheng Chen\textsuperscript{2}, Chu-Song Chen\textsuperscript{3}, Shao-Yi Chien\textsuperscript{1}\\ 
\normalsize \textsuperscript{1}Graduate Institute of Electronics and Engineering, National Taiwan University\\
\normalsize \textsuperscript{2}Research Center for Information Technology Innovation, Academia Sinica \\
\normalsize \textsuperscript{3}Department of Computer Science and Information Engineering, National Taiwan University \\
{\tt \scriptsize jackieliu@media.ee.ntu.edu.tw,}
{\tt \scriptsize pullpull@citi.sinica.edu.tw,}
{\tt \scriptsize chusong@csie.ntu.edu.tw,}
{\tt \scriptsize sychien@ntu.edu.tw}

}

\maketitle

\begin{abstract}
\vspace{-3mm}
Video-based person re-identification (Re-ID) aims at matching the video tracklets 
with cropped video frames for identifying the pedestrians under different cameras. 
However, there exists severe spatial and temporal misalignment for those cropped tracklets due to the imperfect detection and tracking results generated with obsolete methods. 
To address this issue, we present a simple re-Detect and Link (DL) module which can effectively reduce those unexpected noise through applying the deep learning-based detection and tracking on the cropped tracklets. 
Furthermore, we introduce an improved model called Coarse-to-Fine Axial-Attention Network (CF-AAN). 
Based on the typical Non-local Network, we replace the non-local module with three 1-D position-sensitive axial attentions, in addition to our proposed coarse-to-fine structure. 
With the developed CF-AAN, compared to the original non-local operation, we can not only significantly reduce the computation cost but also obtain the state-of-the-art performance ($\mathbf{91.3}\%$ in rank-1 and $\mathbf{86.5}\%$ in mAP) on the large-scale MARS dataset.
Meanwhile, by simply adopting our DL module for data alignment, to our surprise, 
several baseline models can achieve better or comparable results with the current state-of-the-arts.
Besides, we discover the errors not only for the identity labels of tracklets but also for the evaluation protocol for the test data of MARS. We hope that our work can help the community for the further development of invariant representation without the hassle of the spatial and temporal alignment and dataset noise. The code, corrected labels, evaluation protocol, and the aligned data will be available at \small \url{https://github.com/jackie840129/CF-AAN}.
\vspace{-5mm}
\end{abstract}
\vspace{-2mm}
\section{Introduction}
\vspace{-1mm}
\label{sec:intro}
Person re-identification (Re-ID) aims to solve the problem of identifying pedestrians in a multi-camera surveillance system. 
Many researches focus on the image-based setting that identifies people with still images~\cite{parsing,hacnn,pcb,dgnet,relation}. 
Recently, video-based Re-ID~\cite{ap3d,tclnet,vrstc,nvan} has drawn significant attention since comparing continuous video sequences is more practical for the real-world scenarios.
Besides, the appearance information with spatial and temporal relations in a video tracklet contains more cues for matching people under different views. 
The most commonly used methods for tackling video sequences are the 3D convolution~\cite{3dconv} and Non-local operation~\cite{non-local}, which can effectively aggregate the features along the spatial and temporal dimensions.  However, in contrast to image-based setting that the training and testing images of pedestrians are chosen with the least noise from their belonged tracklets, the video-based Re-ID faces more unexpected challenges owing to the imperfect bounding box detection. 

\begin{figure}[t]
	\centering
    \includegraphics[width=0.7\linewidth]{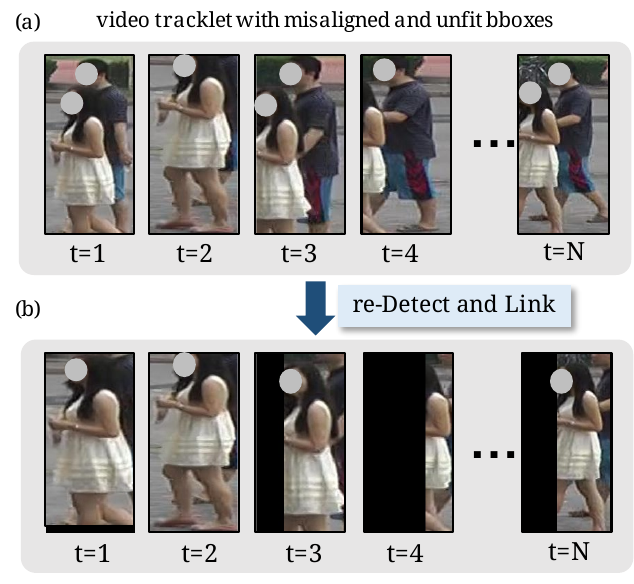}
    \mycaption{Video tracklet processed with our re-Detect and Link (DL) module}{(a) The tracklet is with unexpected noise, where the frame at $t=4$ is dominated by the man in blue shirt, but the ground truth identity is the girl in white dress. (b) The tracklet after DL is less interfered by the man
    .}
    \label{fig:fig1}
    \vspace{-6mm}
\end{figure}

MARS~\cite{mars}, the largest video-based Re-ID dataset so far, adopted traditional DPM~\cite{dpm} as the pedestrian detector and applied GMMCP tracker~\cite{gmmcp} with color histogram as image features, which is not robust enough for linking people under a complicated environment with occlusion. 
As Fig.~\ref{fig:fig1}(a) illustrated, the bounding boxes generated by the weak detector cannot well fit the desired identity (the girl with white dress). Recently, Gu~\etal~\cite{ap3d} proposed the appearance preserving module (APM) inserted before the 3D convolution to align the features along the temporal axis based on each anchor (the center) frame of the 3D sliding windows. 
Although the method achieves the state-of-the-art performance, it still cannot resolve the problems when the center frame contains unexpected noise, such as the fourth frame in Fig~\ref{fig:fig1}(a), where the APM will align the third and fifth frames (if the filter size along the temporal axis is 3) according to the appearance of the man with blue T-shirt.

Since efficient deep-learning algorithms are well-developed for object detection and tracking in the past few years~\cite{fasterrcnn,ssd,yolov4,towards}, to help the community for the further development of invariant representation without the hassle of the spatial and temporal alignment, we revised the original dataset with our proposed simple but effective \textbf{\textit{re-Detect and Link}} (\textbf{DL}) module.
Because we cannot obtain the original video stream containing the whole image frame, our DL module serves as a pre-processing technique on the Re-ID data. Given the original noisy cropped sequence, we first apply a pretrained efficient object detector~\cite{yolov4} to generate much tighter bounding boxes. If there are multiple pedestrian candidates, we will link the pedestrians based on their image features using ID-discriminative embedding (IDE)~\cite{ppp}. Last, according to the aspect ratio and the position of the bounding box, we resize and pad it to the desired image size, as shown in Fig~\ref{fig:fig1}(b). 
Surprisingly, with only the input data processed by our DL module first, even the C2D baseline method~\cite{ap3d}, which only averages the features of each image generated by 2D ResNet~\cite{resnet}, or the normal 3D convolution model P3D-C~\cite{p3d} can achieve promising results. As shown in Table~\ref{tab:intro}, we conduct more experiments on the original and the processed data using some recent state-of-the-arts reproduced by ourselves. From the table, it can be seen that originally the AP3D with the APM module proposed by Gu~\etal~\cite{ap3d} (the last row) can boost about $2\%$ in mAP compared to its P3D-C counterpart (the second row). However, with the aligned input images generated by our DL module, it only increase $0.4\%$ in mAP. This shows that the state-of-the-art AP3D cannot extract more discriminative features for Re-ID given the already aligned data.
Furthermore, we can see that the self-attention based Non-local Network~\cite{ap3d,nvan} combined with our DL module can achieve the new state-of-the-arts, which means the self-attention on the less noisy data can generate more representative Re-ID features.
Thus, in the next step, we focus on the Non-local Network but developing an efficient baseline model which can perform comparable results.

Non-local Network achieves state-of-the-art performance on video-based Re-ID, but its high computation cost remains an issue for practical usage.
Each feature point along spatial and temporal dimensions needs to compute its self-attention map for all other points. 
To reduce the computation while retaining the performance of Non-local Network on Re-ID, following the idea of axial-attention~\cite{axial,ccnet} and the multi-granularity (coarse-to-fine) structure in~\cite{gran}, we propose the \textit{Coarse-to-Fine Axial-Attention Network} (CF-AAN). With the axial attention, we can factorize the 3D attention operation into three 1-D attention ones sequentially along the height-, width- and temporal-axis.
To further boost the efficiency, in contrast to~\cite{gran} that adding the coarse-to-fine module after the whole model backbone, we directly integrate it into our axial-attention. We split the input tensor into multiple scales along the channel dimension, and transform the spatial dimension from coarse to fine scales. 
To the best of our knowledge, we are the first to adopt axial-attention in video-based Re-ID. Our DL+CF-AAN approach not only achieves the state-of-the-art performance on two large-scale datasets~\cite{mars,dukev}, but also significantly save the computation as compared with vanilla Non-local Network, which can be regarded as an efficient baseline self-attention method.

\begin{table}[t]
    \centering
    \caption{Performance of recent state-of-the-arts reproduced with our re-Detect and Link (DL) on MARS~\cite{mars}. The score with underline is the runner-up.}
    \label{tab:intro}
    \vspace{-2mm}
    \scalebox{0.85}{
    \begin{tabular}{l|cc|cc}
    \hline
    \multirow{2}{*}{Method} & \multicolumn{2}{c|}{ Original Results } & \multicolumn{2}{c}{\textbf{w/ our DL}}  \\ 
    \cline{2-5}
                & mAP   & rank-1  & mAP  & rank-1 \\ \hline \hline 
    FT-WFT~\cite{ft-wft}  & 82.9 & 88.6  & 83.8 & 90.0 \\
    P3D-C~\cite{p3d,ap3d}  & 83.1 & 88.5  & 85.0 & 91.0 \\
    C2D~\cite{ap3d}  & 83.4 & 88.9  & 84.9 & 91.0 \\
    Non-Local~\cite{ap3d,nvan}  & 85.0 & 89.6 &  \bf{86.2} & \bf{91.4} \\
    TCLNet~\cite{tclnet}  & \underline{85.1} & \underline{89.8}  & \underline{85.8} & 90.8 \\
    AP3D~\cite{ap3d}  & \bf{85.1} & \bf{90.1}  & 85.4 & \underline{91.0}\\
    \hline
    \end{tabular}}
    \vspace{-2mm}
\end{table}

\begin{figure}[t]
	\centering
    \includegraphics[width=0.9\linewidth]{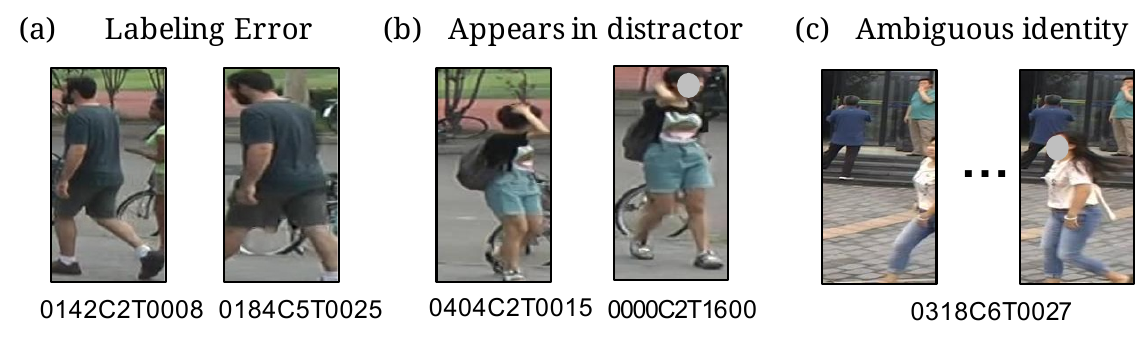}
    \vspace{-3mm}
    \caption{Illustration of the labeling errors and ambiguous cases in MARS~\cite{mars} testing set. More samples and details can be found in Sec.~\ref{sec:label_cleaning}.}
    \label{fig:introerror}
    \vspace{-5mm}
\end{figure}

In addition to the application of our DL module that can significantly improve the performance, we also find that there are multiple \textbf{labeling errors or noises} in the MARS testing data. As shown in Fig.~\ref{fig:introerror}(a), the two tracklets are labeled as different identities (ID 142 and 184) but are actually the same person. Or in Fig.~\ref{fig:introerror}(b), the tracklet with ID 404 in camera 2 also appears in the distractor class (ID 0), which will make the model easily match the two tracklets but counted as an error matching in the evaluation. 
There are also some ambiguous cases that cannot be distinguished even by human. As in Fig~\ref{fig:introerror}(c), the ID 318 is the man in blue behind but the bounding boxes also contains the woman in white (ID 322). Thus, we revise the labels in the testing set and the original evaluation protocol. The details will be described in Sec.~\ref{sec:label_cleaning}. We hope that the release of our DL processed test data 
on MARS can help the community to validate their methods on a clean testing set and push the further development of improved representation. 


Our contributions can be highlighted as follows:
\begin{compactitem}
\item We propose a re-Detect and Link module that can align the noisy tracklet on the image level, which makes a simple method achieving comparable performance.
\item Besides the aligned data, we additionally provide revised identity labels and evaluation protocol in MARS testing set, which helps validate the new methods on a corrected benchmark.
\item A baseline Coarse-to-Fine Axial Attention Network (CF-ANN) is proposed, which performs axial-attention from coarse to fine levels, which not only reduces the computation cost but achieves the promising performance.
\end{compactitem}


\vspace{-1mm}
\section{Related Work}
\vspace{-1mm}
We briefly review the recent related development of video-based person re-identification as follows.
\vspace{-4mm}
\paragraph{Video-based Person Re-identification}
Compared to image-based Re-ID, video-based setting contains more frames and additional temporal information.
Typically, researchers aggregate the information among a tracklet with temporal modeling or attention-based algorithms and optimize the model with discriminative learning~\cite{ppp} or metric learning~\cite{triplet}. For temporal modeling, McLaughlin~\etal~\cite{rnnreid} first applied Recurrent Neural Network (RNN) on the frame-wise CNN features to allow information flowing among different frames and obtain a sequence-level representation. Inspired by the success of 3D Convolutional Neural Network on action recognition~\cite{action,3dact}, the work~\cite{m3d} first adopted the 3D convolution to automatically learn the relation from low- to high-level features along spatial and temporal dimensions. In order to resolve the alignment problems, Gu~\etal~\cite{ap3d} then proposed an APM module inserted before the 3D convolution to align the features among each 3D filters. In contrast to treating each frame even, some works utilized the attention mechanism that can focus on some specific regions representing the identity better~\cite{drsa,duatm,snippet,sta,nvan,gran}.  Li~\etal~\cite{drsa} proposed multiple spatial attention modules that can focus on many important spatial regions across different frames and the spatial features are then aggregated by a learnable temporal attention. Chen~\etal~\cite{snippet} adopted a novel co-attention mechanism that can dynamically learn the feature representation based on the query and gallery pairs. Zhang~\etal~\cite{gran} explored the attention mechanism with a global reference, which can effectively learn the attention more on the region with close relation to the global guidance. Besides performing attention on the last layer of CNN features, Liu~\etal~\cite{nvan} started to aggregate the popular non-local self-attention~\cite{non-local} inside the CNN backbone. Compared to those methods, our model is based on the self-attention operation and added with computation efficient structures into the model design.
\\
\textbf{Self-Attention}~~
Since the self-attention based Transformer~\cite{transformer} obtained a great success in nature language processing, recently many works started to tackle the problems in computer vision with self-attention~\cite{non-local,aacn,1616,stand,axial,axialdeeplab,detr,ccnet}. The plain type of the self-attention is the non-local network~\cite{non-local} without the position encoding and multi-head attention and was proposed to solve the problem of video classification. Because the non-local self-attention is computation demanding, axial-attention~\cite{axial,axialdeeplab} were proposed to factorize the operation into multiple 1-D self-attentions, which can extremely reduce the cost. Dosovitskiy~\etal~\cite{1616} and Carion~\etal~\cite{detr} even integrated the whole transformer respectively into the image classification and object detection tasks, and they all obtained comparable performance to the methods with original CNN backbone. Our work focus on adding the efficient axial-attention module with our proposed coarse-to-fine structure into typical CNN to learn spatially and temporally attentive feature representation. 
\\
\textbf{Dataset and Evaluation Protocol Revision}~~
In the field of person re-identification, there is no work exploring and revising the original imperfect data or discussing the evaluation protocols, labeling errors, and the ambiguous cases in the testing set. We found that in the field of face detection, there are some works investigating the noise in the labels or the bias in evaluation protocols~\cite{facebw,proximity,global-local}. Mathisas~\etal~\cite{facebw} provided improved annotations of existing face datasets and evaluation criteria that resolved the original problems. Besides, they also showed that when properly used, a simple vanilla baseline can reach top performance on face detection. Lin~\etal~\cite{proximity} and Zhang~\etal~\cite{global-local} both tried to remove the data with labeling errors before training by utilizing the inherent data distributions. 
Compared to our work, we adopt pretrained deep learning-based object detector to refine the original test data that are unfit to the target identity. With the aligned data, even a simple baseline method can achieve outstanding performance. Moreover, we manually check the errors with the existing Re-ID evaluation protocol and provide some revision of not only the labels but the evaluation protocol.  
\vspace{-3mm}


\begin{figure*}[t!]
	\centering
    \includegraphics[width=\linewidth]{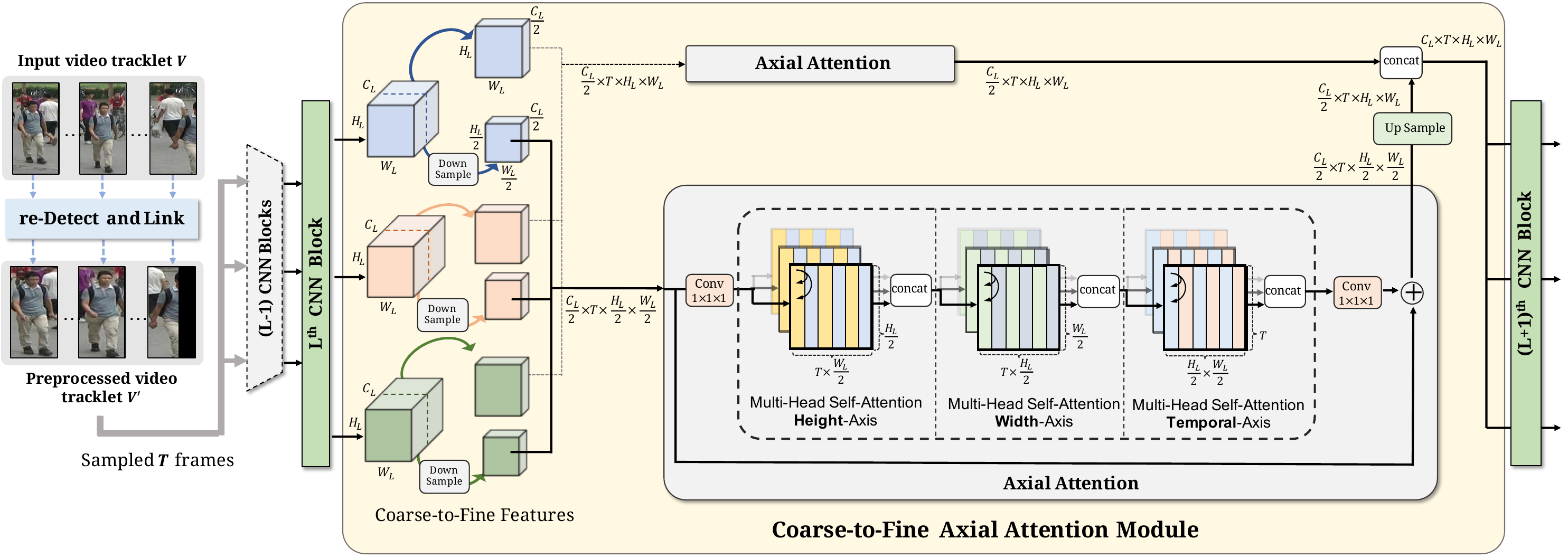}
    \mycaption{Pipeline of our DL and CF-AAN architecture}{The original tracklet $\mathcal{V}$ is first fed into the DL module and become the processed tracklet $\mathcal{V}^\prime$, which will then be sampled and fed to CF-AAN. We demonstrate one CF-AA module between the $L^{th}$ and $(L+1)^{th}$ CNN block. There are two scales of features and the axial-attention will perform on each of them. The outputs will be up-sampled and concatenated to become the input of the next CNN block.}
    \label{fig:big}
    \vspace{-2mm}
\end{figure*}

\section{The Proposed Method}
\vspace{-1mm}
Fig.~\ref{fig:big} demonstrates the pipeline of our re-Detect and Link (DL) techniques and the proposed Coarse-to-Fine Axial Attention Network (CF-AAN). Given an original imperfect video tracklet $\mathcal{V}$ with $N$ images, $\mathcal{V}=\{I_1,I_2,..,I_N\}$, we first adopt our DL module to obtain the processed tracklet $\mathcal{V}^\prime$, which is more robust and aligned. The detail of our DL will be described in Sec.~\ref{sec:DL}. Then, as the typical pipeline of video-based Re-ID, we sample $T$ frames from $\mathcal{V}^{\prime}$ as the input of our CF-AAN. Our network consists of a backbone CNN and multiple Coarse-to-Fine Axial Attention (CF-AA) modules, which are separately inserted between the CNN blocks. The operations in our CF-AAN are described in Sec.~\ref{sec:CFAAN}. Last, in Sec.~\ref{sec:optimize}, the video features generated by our CF-AAN will be aggregated with the masks created along with DL module and optimized with the common losses for Re-ID.
\vspace{-1mm}
\subsection{Data Alignment with re-Detect and Link}
\label{sec:DL}
\begin{figure}[t]
	\centering
    \includegraphics[width=0.9\linewidth]{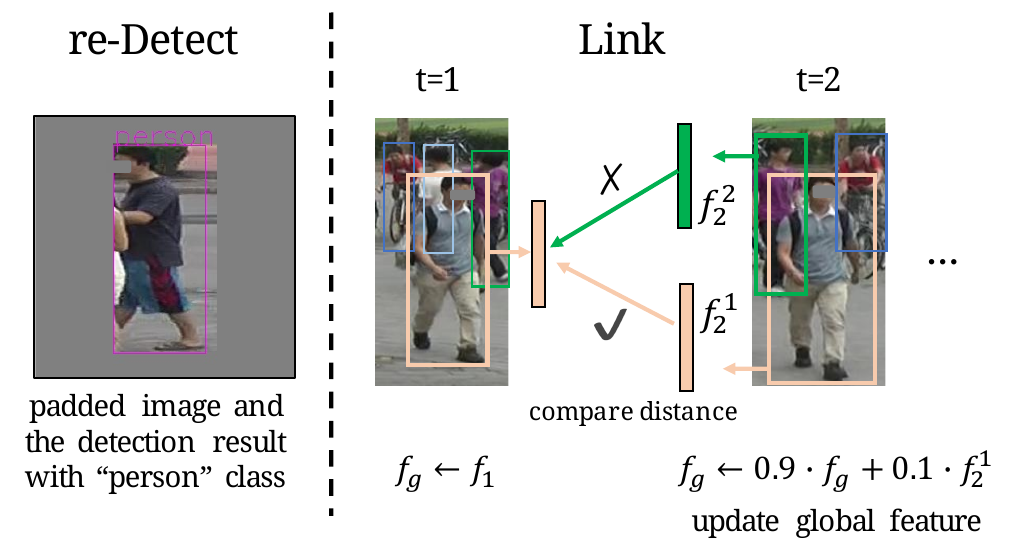}
    \vspace{-2mm}
    \caption{Illustration of the re-Detect and Link module.}
    \label{fig:dl}
    \vspace{-4mm}
\end{figure}

\vspace{-1mm}
With the noisy video tracklet $\mathcal{V}$ with $N$ images, we sequentially perform our re-Detect and Link (DL) method on each video frame and create a new processed tracklet $\mathcal{V}^\prime$ with $N$ frames, too. As illustrated in Fig.~\ref{fig:dl}, first, all images are padded and fed to the object detector~\cite{yolov4} to generate candidate bounding boxes with the ``person'' class. For the first frame, if there are multiple candidates, we will assume that the bounding box with larger area is the desired one. Then, similar to the feature-based real-time object tracking~\cite{towards}, we extract the feature $f$ of the cropped image $I^\prime_{1}$ by the IDE feature extractor trained on the original dataset~\cite{mars}, and save it as the global feature $f_g=f_1$. Next, for each consecutive frame $i$, if there are multiple candidates, we will compare each extracted feature $f^{j}_{i}$ to the global feature $f_g$ and choose the one with the smallest Euclidean distance, where $j$ is the index of the candidate bounding box in $i^{th}$ frame. After choosing the candidate for the $i^{th}$ frame, the global feature will then be updated by 
\begin{equation}
    \vspace{-2mm}
    f_g = \alpha f_g + (1-\alpha) f_i ~,
\end{equation}
where $\alpha$ is set to $0.9$ in our case. 

Note that in Re-ID datasets, we cannot obtain the original full image frame captured by cameras and perform our DL method. Thus, after we apply object detection on the noisy cropped image, we may obtain a new cropped identity with only part of his/her appearance, as shown in Fig~\ref{fig:fig1}(b).
According to the aspect ratio and the position of the bounding box in the image, if the bounding box is slim (the height is much larger than the width) and its position is on the left (right) of the image, we will shift it to the right (left), resize it based on its original aspect ratio and pad it to the desired image size. Furthermore, we also create a mask $M_i$ of the output image $I_i^{\prime}$ representing whether each pixel is the padded one or not. This mask will then be applied in the feature aggregation of our CF-AAN, which will be described in Sec.~\ref{sec:optimize}.
\\
\textbf{Discussion}~~~Comparing to other methods proposing an automatically learned feature alignment mechanism inside their backbone model~\cite{align,ap3d}, our DL module adopts an additional object detector to help reduce the original noise in the data. It seems that our method requires additional computation cost but and utilizes extra information. However, we want to point out that the goal behind our DL module is to simulate a nowadays real-life scenario with efficient and robust deep learning-based object detection and tracking \textbf{before} Re-ID. Thus, when it really comes to the Re-ID phase, actually there has been no need for this additional cost of DL module on the input tracklet. Furthermore, as shown in the Table~\ref{tab:intro}, with the aligned data, the simplest baseline can obtain a promising Re-ID result and the original state-of-the-art methods that specifically deal with the problems of misalignment will not retain its competitiveness. We think that with the release of the data processed by our simple alignment method, it can help the community explore more on the attention-based methods or the methods for learning invariant feature representation.

\vspace{-1mm}
\subsection{Coarse-to-Fine Axial-Attention Network}
\vspace{-1mm}
\label{sec:CFAAN}
As shown in Table~\ref{tab:intro}, the existing self-attention based Non-local Network can achieve the best result on the aligned data. However, the efficiency is the main drawback. We propose a simple method called  Coarse-to-Fine Axial-Attention Network that contains a coarse-to-fine mechanism and a position-sensitive axial-attention which dramatically reduce the computation burden but retain comparable performance. 

\vspace{-4mm}
\paragraph{Self-Attention:} We first introduce the typical 3D self-attention~\cite{non-local} operation as follows. Given an input feature map $x \in \mathbb{R}^{C_{in}\times T \times H \times W }$ with channels $C_{in}$, temporal length $T$, height $H$, and width $W$, the output $y$ at position $o=(i,j,t)$, $y_o \in \mathbb{R}^{C_{out}}$, is computed by aggregating all the projected input as :
\begin{equation}
    \vspace{-2mm}
    y_{o} = \sum_{p\in \mathcal{N}}{\text{softmax}_p(q^T_o k_p)v_p}
\end{equation}
where $\mathcal{N}$ is the set of the whole $HWT$ locations, and queries $q_o$, keys $k_o$, and values $v_o$ are three different linear projections of the input $x_o, ~\forall o \in \mathcal{N}$ from dimension $C_{in}$ to intermediate $C_{q,k}$ for query and key projection or $C_{out}$ for value projection. As opposed to convolution which only captures local relations, this mechanism allows us to capture related but non-local context in the whole feature map. Commonly, it will be inserted into multiple locations between the backbone CNN layers, and each complexity is $\mathcal{O}(H^2W^2T^2)$. 

%

\vspace{-4mm}
\paragraph{Axial-Attention:} To reduce the computation of non-local self-attention, in 2D image classification tasks, the axial-attention has been proposed~\cite{axial}, they factorized the 2D self-attention operation into two 1-D axial-attentions. When applied to our video-based Re-ID, the 3D self-attention will be consecutively factorized into height-axis, width-axis and the temporal-axis. With this transformation, the complexity can be reduced to $\mathcal{O}(H^2WT+HW^2T+HWT^2)$. The formulation of the axial-attention, with the height-axis as an example, is as follows.
\begin{equation}
    y_{o} = \sum_{p\in \mathcal{N}_{H \times 1 \times 1}}{\text{softmax}_p(q^T_o k_p)v_p}
\end{equation}
where the location $p$ only lies along the $H$ axis.

Furthermore, based on the concept proposed in the Transformer~\cite{transformer}, many works start to encode the positional encoding into the self-attention structure~\cite{detr,aacn,stand}. Thus, the final method we adopt is based on the positional-sensitive axial-attention proposed in~\cite{axialdeeplab}, where the 
learnable positional encoding vectors depends on the query vectors, key vectors and the value vectors. 
The formulation is as follows with the height-axis as an example.
\begin{multline}
y_{o} = \sum_{p\in \mathcal{N}_{H \times 1 \times 1}}{\text{softmax}_p(q^T_o k_p + q^T_or^q_{p-o} + k^T_pr^k_{p-o})}\\(v_p+r^v_{p-o})
\vspace{-2mm}
\end{multline}
where the $r^{q}_{p-o}$, $r^{k}_{p-o}$, and $r^{v}_{p-o}$ are the learned relative positional embedding. 
Besides, in practice, as shown in Fig~\ref{fig:big}, we will extend the single-head attention into multi-head attention to generate a mixture of affinities. To retain the complexity, if there are $M$ parallel single-head attentions, in the $m^{th}$ head, each dimension of the $q^m$,$k^m$, and $v^m$ will be shrunk to $\frac{C_{q,v}}{M}$ and $\frac{C_{out}}{M}$. The dimension of the learnable positional vectors $r^q_{p-o}$, $r^k_{p-o}$ and $r^v_{p-o}$ are also shrunk but the vectors are shared across each head. Thus, the final output $z_o$ will be the concatenation of each head, $z_o=concat_m(y_o^m)$, with the same dimension $C_{out}$. 
Last, after conducting the axial-attention (AA) along the three dimensions, we will project the output feature from dimension $C_{out}$ back to $C_{in}$ and added with the input tensor $x$ to become a new refined tensor $x^\prime$, which is formulated as follows.
\begin{equation}
    x^\prime = x + Conv(\text{AA}^{T}(\text{AA}^{W}(\text{AA}^{H}(x))))
\end{equation}

\vspace{-4mm}
\paragraph{Coarse-to-Fine Axial Attention:} In addition to multi-head attention that learns different structure of affinities, we propose a Coarse-to-Fine Axial-Attention module (CF-AA) that not only makes the self-attention learn on different scales of the spatial dimension but further reduce the computation. Different from~\cite{gran}, which can only perform multi-scale structure on the last layer of CNN backbone with the smallest resolution, we can apply our structure along with the axial-attention from the mid-level stage to high-level stage inside the backbone. As shown in Fig.~\ref{fig:big}, we split the input tensor $x$ with $S$ scales along the channel dimension and for the $s^{th}$ scale, we downsample the spatial resolution to $H_s\times W_s$, where $H_s=\frac{H}{2^{s-1}}$ and  $W_s=\frac{W}{2^{s-1}}$. Thus, if $S=2$ as an example, the original input tensor $x$ will be split into $x_1 \in \mathbb{R}^{\frac{C_{in}}{2}\times T \times H \times W}$ with a fine scale and $x_2 \in \mathbb{R}^{\frac{C_{in}}{2}\times T \times \frac{H}{2} \times \frac{W}{2}}$ with a coarse scale. The split tensors are then separately fed into the axial-attention and the outputs are upsampled and concatenated along the channel dimension in order to retain the original tensor size.

\vspace{-1mm}
\subsection{Feature Aggregation and Optimization}
\vspace{-1mm}
\label{sec:optimize}
Our CF-AAN contains a 2D CNN backbone and several CF-AA modules inserted between the CNN blocks. After the last CNN layer, there will be $T$ tensors with size $\mathbb{R}^{C^\prime \times H^\prime \times W^\prime}$. As mentioned in Sec.~\ref{sec:DL}, because there are some input pixels which are the padded ones without any information, we first downsample the mask $M$ to $M^\prime$ according to the spatial dimension $H^\prime$ and $W^\prime$
, and utilize the mask to average-pool on the desired spatial region to generate $T$ vectors with $C^\prime$ dimension. Then, we aggregate the features with the typical average operation followed by a Batch-Normalization (BN) layer~\cite{bn} to create the final feature representation $f_\mathcal{V}$ of the video tracklet.
To optimize the network, we follow the two loss combinations in BoT~\cite{bot}, which consists of a batch-hard triplet loss~\cite{triplet} on the features before BN and a cross-entropy loss~\cite{ppp} on the identity classifier (a fully-connected layer) after the feature $f_\mathcal{V}$.
\begin{table*}[t!]
    \centering
    \mycaption{The Ablation Study of our DL and CF-AAN}{We compare the effectiveness of our DL and all the components in CF-AAN with the computation cost (GFLOPs) and performance on MARS. Except the baseline itself, all other computation costs are the increase comparing to the baseline method. $C_B$: the computation cost of the baseline method. }
    \label{tab:ablation}
    \vspace{-1mm}
    \scalebox{0.9}{
    \begin{tabular}{l|c|cccc|c|cc}
    \hline
    \multirow{2}{*}{Method} & \multirow{2}{*}{w/ our DL} & \multicolumn{4}{c|}{Self-attention Module}  & \multirow{2}{*}{\bf{\#GFLOPs}} & \multicolumn{2}{c}{\bf{MARS}} \\ 
    \cline{3-6}\cline{8-9}
    & & Self-attention & \#~of heads & Posi. Encoding & \# of scales & & mAP & R-1 \\ 
    \hline \hline
    \multirow{2}{*}{Baseline}  & \xmark & \xmark&\xmark &\xmark&\xmark & \multirow{2}{*}{24.520~ ($C_B$)} & 83.4 & 87.7 \\
                               & \cmark & \xmark&\xmark &\xmark&\xmark &                         & 85.1 & 89.7 \\
    \hline
    Non-local  & \cmark & 3D self-attention & 1 &\xmark & 1 & $C_B$+17.213 & 86.2 & \bf{91.4} \\
    \hline
    \multirow{6}{*}{Axial-based}    & \cmark & Axial-attention & 1 & \xmark  & 1 & $C_B$+0.361& 86.0 &91.1 \\ 
                   & \cmark & Axial-attention & 8 & \xmark  & 1 & $C_B$+0.361& 86.2 &91.2 \\ 
                    & \cmark & Axial-attention & 8 & Sinusoidal  & 1 & $C_B$+0.377 & 86.0 &91.1 \\ 
                    & \cmark & Axial-attention & 8 & Relative & 1& $C_B$+0.424 & 86.4 &91.2 \\ 
                   \cline{2-9}
                   & \cmark & Axial-attention & 8 & Relative & 2& $C_B$+0.245  & 86.4 &91.3 \\ 
                   & \cmark & \bf{Axial-attention} & \bf{8} & \bf{Relative} & \bf{4} & \bf{$C_B$+0.126}  & \bf{86.5} &91.3 \\ 
    \hline
    \end{tabular}}
    \vspace{-1mm}
\end{table*}

\vspace{-3mm}
\section{Experiments}
\vspace{-1mm}
In this Section, we conduct extensive evaluation and ablation studies of the proposed approach in addition to the analysis and correction of data noise and labeling errors for the evaluation dataset.
\vspace{-2mm}
\subsection{Datasets and Evaluation Protocol.}
\vspace{-1mm}
We evaluate the proposed method on two large-scale datasets, MARS~\cite{mars} and DukeMTMC-VideoReID~\cite{dukev}, abbreviated as DukeV. MARS consists of 17,503 tracks and 1,261 identities. Each track has 59 frames on average. Deformable Part Model~\cite{dpm} is employed to detect pedestrians and GMMCP~\cite{gmmcp} is used to track pedestrians. To make the MARS dataset even more challenging, they include 3,248 distractor tracks. For DukeV, it comprises 4,832 tracks and 1,404 identities. Each track contains 168 frames on average. Different from MARS, the detection and tracking ground truth are manually labeled.
We use the rank-1 (R1) in the Cumulative Matching Characteristics (CMC) and the mean Average Precision (mAP)~\cite{market} as evaluation metrics.

\vspace{-2mm}
\subsection{Implementation Details.}
\vspace{-1mm}
\paragraph{re-Detect and Link.} Our object detector is the YoloV4~\cite{yolov4} pretrained on the COCO dataset~\cite{coco}. The IDE~\cite{mars} model for linking the candidates is a ResNet-50~\cite{resnet}. We perform our DL module both on MARS and DukeV dataset. However, because only the MARS dataset is adopted with traditional detector and tracker, where the data in DukeV is manually labeled, the processed data of DukeV is almost the same as before. \\
\textbf{CF-AAN.}~~For our CF-AAN, we adopt ImageNet pre-trained ResNet-50~\cite{resnet} as our backbone. Similar to the structure of Non-local Network~\cite{non-local}, we insert 5 CF-AA modules, 2 after $conv3\_3$, $conv3\_4$ and another 3 after $conv4\_4$, $conv4\_5$, and $conv4\_6$ respectively. In our coarse-to-fine structure, we split the feature into four levels ($S=4$) and in each axial-attention, we set the number of head $M=2$. Thus, the total number of heads in a coarse-to-fine axial-attention module is equals to $8$, which is similar to the original axial-attention network~\cite{axialdeeplab}. In the training stage, we sample $T=6$ images as an input tracklet. Each frame in a tracklet is resized to $256\times 128$ and synchronously augmented with random horizontal flip. As for the optimizer, Adam~\cite{adam} with weight decay $5\times{10}^{-5}$ is adopted. We train the model for $220$ epochs. The learning rate is initialized to ${10}^{-4}$ and multiplied by $0.1$ after every $50$ epochs. In the testing stage, for each tracklet, we split it into several 6-frame clips, and then the feature representations for each clip are averaged to become the final representation.

\begin{figure}[t]
	\centering
    \includegraphics[width=0.9\linewidth]{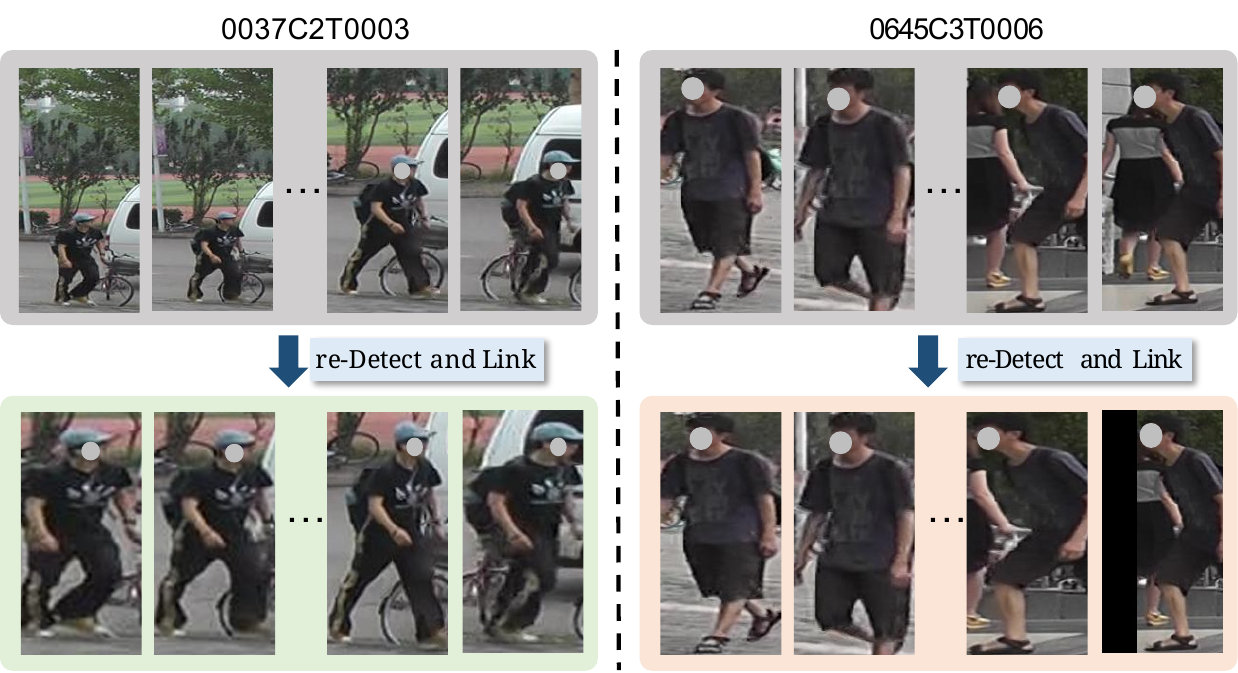}
    \caption{Examples of video tracklets processed by our DL.}
    \label{fig:dl_vis}
    \vspace{-5mm}
\end{figure}

\vspace{-2mm}
\subsection{Ablation Study}
\vspace{-1mm}
In Table~\ref{tab:ablation}, we conduct ablation study on our proposed re-Detect and Link (DL) module and our Course-to-Fine Axial-Attention Network (CF-AAN). Besides the Re-ID performance, we also calculate the computation cost of inference in terms of GFLOPs. 
We first analyze the effectiveness of the DL module on our baseline method (the first two rows). Our ``Baseline'' method, with $24.52$ GFLOPs operations, contains the same ResNet-50 backbone, types of losses and training details but without all the axial-attention modules, which is just the average of features in each frame. We can clearly see that with the aligned data processed by DL, there is an obvious improvement of the performance ($1.7\%$ in mAP). Thus, the alignment of the input video tracklet is crucial and important for the subsequent feature extraction. We also demonstrate some extra examples in Fig.~\ref{fig:dl_vis}. We can see that the problems of misalignment in the left tracklet and the multiple candidates in the right tracklet are resolved after processed with the DL module.  

Next, we compare the self-attention based methods. The first one is Non-local Network (the $3^{th}$ row), which is with single head 3D self-attention but without the positional encoding. Although it can improve about $1.1\%$ in mAP compared to the baseline, the computation also increases ($\mathbf{+17.213}$ GFLOPs), which is extremely large and almost equal to the baseline. After replacing the operation with axial-attention, the computation can reduce to only $+0.361$ GFLOPs, while the performance slightly decrease owing to its factorized self-attentions. With the multi-head structure (the $5^{th}$ row), it can retain the computation cost but increase the performance.

We then apply two types of positional encoding to explore their effectiveness. The first one is the sinusoidal encoding (the $6^{th}$ row) which is the same as the experiments in~\cite{aacn} and the learnable relative positional embedding (the $7^{th}$ row) proposed in~\cite{axialdeeplab}. We can see that there is no significant influence of all kinds of positional encoding but the relative and learnable characteristics are the best for Re-ID, which can achieve $86.4\%$ in mAP. Last, in the last two rows, we demonstrate the benefits brought by our coarse-to-fine structure. We can see that, because the spatial dimensions decrease in the coarser scale, the total operations also decrease. When the number of scales is 4, the operation can increases only $0.126$ GFLOPs compared with the baseline, which is only about $\mathbf{1\%}$ of those in Non-local Network. Furthermore, owing to the coarse-to-fine structure that makes the self-attention learn on different scales, the performance even increases to $\mathbf{86.5}\%$ in mAP on MARS dataset. The CF-AAN with four scales is our final model performing the video-based Re-ID.

\vspace{-1mm}
\subsection{Comparison with State-of-the-art Approaches}
\vspace{-1mm}
We compare recent state-of-the-art approaches with our methods on MARS and DukeV datasets in Table~\ref{tab:sota}. We can see that in the past, the methods that globally perform attention mechanism on the last CNN features are the mainstream for dealing with video tracklet~\cite{drsa,duatm,snippet,sta}. However, the noise and unaligned appearance between frames make it hard to learn a robust attention score. In another way, TCLNet~\cite{tclnet} conduct the attention frame by frame, which is less interfered by the alignment problems. AP3D~\cite{ap3d} is the recent work that adopts 3D convolution with a feature alignment module inserted between 3D CNN blocks. We can see that once reducing this unaligned problem, a 3D CNN can achieve the best results (in R-1). The MG-RAFA~\cite{gran} is also the attention-based method, but they adopt the multi-granularity (multi-scales) structure on the output of the CNN features, where the features will then be fed to their global attention methods. This structure obtains the best results in mAP. Our method consists of a simple but effective pre-processing DL module followed by an extremely efficient CF-AAN. Different from~\cite{gran}, our coarse-to-fine structure is inserted with the axial-attention module between the backbone CNN blocks. We can see that our methods achieve promising performance, which outperform AP3D~\cite{ap3d} $\mathbf{1.4\%}$ in mAP and $\mathbf{1.2\%}$ in R-1 on the MARS dataset. Although the data in DukeV are manually labeled, our model still can retain comparable performance. Thus, in summary, with almost no extra computation cost compared to the baseline, where conducting the DL module is also effortless in real-life scenario, we are the state-of-the-art in terms of the popular mAP metric for the video-based person Re-ID task.

\begin{table}[t]
    \centering
    \mycaption{Comparison with state-of-the-arts ($\%$)}{The score with underline is the runner-up.}
    \label{tab:sota}
    \vspace{-2mm}
    \scalebox{0.87}{
    \begin{tabular}{l|cc|cc}
    \hline
    \multirow{2}{*}{Method} & \multicolumn{2}{c|}{MARS}& \multicolumn{2}{c}{DukeV} \\ 
    \cline{2-5}
            & mAP   & R-1  & mAP   & R-1\\ 
    \hline \hline
    DRSA~~(CVPR18)\cite{drsa} & 65.8 & 82.3 & -& -\\
    EUG~~(CVPR18)\cite{eug}  & 67.4 & 80.8 & 78.3 & 83.6 \\
    DuATM~~(CVPR18)\cite{duatm} & 67.7 & 81.2 & - & - \\
    TKP~~(ICCV19)\cite{drsa} & 73.3 & 84.0 & 91.7 & 94.0 \\
    M3D~~(AAAI19)\cite{m3d} & 74.1& 84.4 & -& -\\
    Snippet~~(CVPR18)\cite{snippet} & 76.1 & 86.3 & - & - \\
    STA~~(AAAI19)\cite{sta} &80.8 & 86.3& 94.9& 96.2 \\
    VRSTC~~(CVPR19)\cite{vrstc} & 82.3 & 88.5 & 93.5 &95.0 \\
    NVAN~~(BMVC19)\cite{nvan} & 82.8 & 90.0 & 94.9 & 96.3 \\
    FT-WFT~~(AAAI20)\cite{ft-wft} &82.9 &88.6 & - & - \\
    TCLNet~~(ECCV20)\cite{tclnet} &85.1 & 89.8 & \underline{96.2} & \bf{96.9} \\
    AP3D~~(ECCV20)\cite{ap3d} & 85.1 & \underline{90.1} & 95.6 & 96.3 \\
    MG-RAFA~(CVPR20)\cite{gran} & \underline{85.9} & 88.8 & - & - \\
    \hline
    DL+CF-AAN~~(\bf{Ours}) & \bf{86.5} & \bf{91.3} & \bf{96.2} & \underline{96.7} \\
    \hline
    \end{tabular}}
    \vspace{-5mm}
\end{table}

\begin{figure*}[t]
	\centering
    \includegraphics[width=0.88\linewidth]{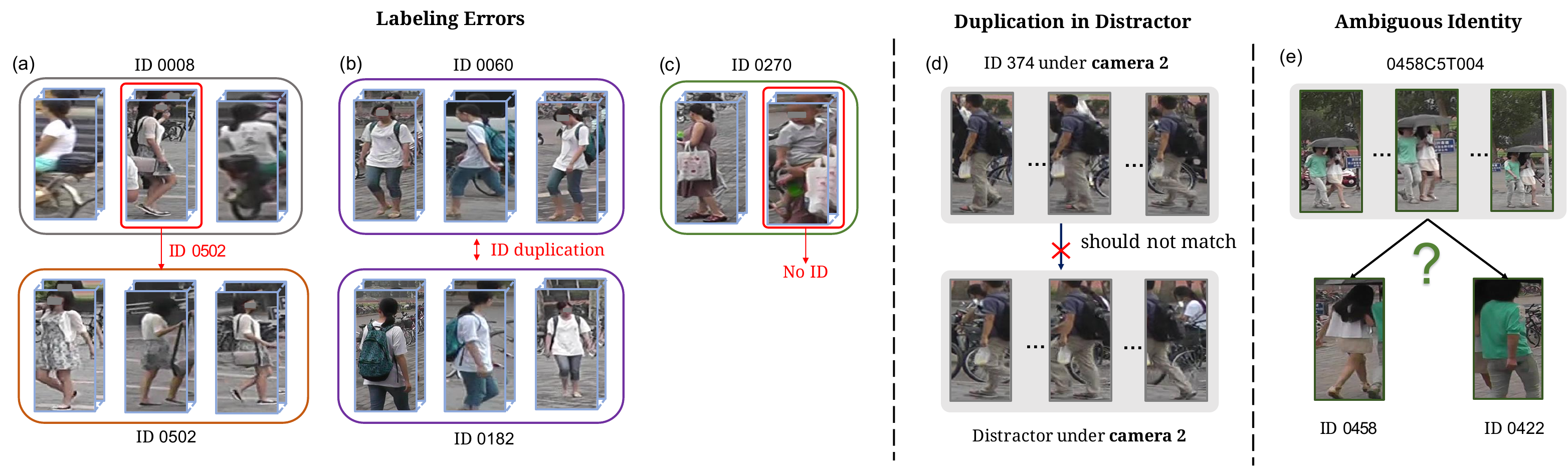}
    \caption{Three kinds of label noises in the MARS testing data.}
    \label{fig:error}
    \vspace{-5mm}
\end{figure*}

\vspace{-2mm}
\subsection{Label Cleaning and New Evaluation Protocols}
\label{sec:label_cleaning}
\vspace{-1mm}
As described in Sec.~\ref{sec:intro}, we found some labeling errors or 
ambiguous cases in the MARS dataset. Thus, we manually check the testing data of the unmatched ones in evaluation and propose a new protocol which additionally address three kinds of new situations: labeling errors, duplication in distractor, and ambiguous identity. \\
\textbf{Purely labeling errors by annotators:}~~
There are also three kinds of labeling errors shown in Figs.~\ref{fig:error}(a)-(c). The first one is that a tracklet may be annotated as another existing identity (\ref{fig:error}(a)). Or, there are completely two groups of tracklets labeled as a different person but in fact with the same identity (\ref{fig:error}(b)). 
Sometimes the tracklet does not belong to any other identities in the testing set. As Fig.~\ref{fig:error}(c) shows, the identity 270 is the woman but the tracklet marked with red box is the baby she holds. For those three cases, we fix the annotation with the correct or new identity.\\
\textbf{Duplication in Distractor Class:}~~
In the original evaluation protocol of MARS~\cite{mars}, if a query tracklet matches a gallery tracklet with the same identity but under the same camera, this match will be ignored because Re-ID aims at matching pairs across cameras. However, the ``distractor class (ID 0)'' in MARS consists of not only the false positive bounding boxes created by pedestrian detector but also some duplicated bounding boxes of the tracklets in testing set. As shown in Fig.~\ref{fig:error}(d), the tracklet with ID 374 under camera 2 will easily match the same tracklet in distractor and strangely counted as an incorrect match. Thus, we revise the evaluation protocol that if a tracklet matches the other one under the same camera with its same identity or the distractor class, they will both be ignored.  \\
\textbf{Ambiguous Identity:}~~
There are some ambiguous cases in the dataset. As the tracklet in Fig.~\ref{fig:error}(e), the unfit bounding box contains two persons (ID 485 and ID 422) from the beginning to the end of the tracklet. With our DL, there is only one person left but the true identity cannot be even distinguished by human. For those cases, we will add an additional ambiguous identity of the tracklet and in the evaluation process, the matches of those identities will all be counted as the correct ones.

Similar to Table~\ref{tab:intro}, we reproduce some existing methods not only with data processed by our DL but evaluated under our new protocols, which are shown in Table~\ref{tab:neweval}. Furthermore, with their released codes, we also demonstrate the computation cost in inference time with fairly 6-frames clip as input data in terms of GFLOPs. We can see that all methods can improve largely by $2.5\%$ in mAP, but our CF-AAN still achieves the best result ($88.9\%$ in mAP). When 
regarding the computation cost, those of our CF-AAN are comparable to the ones of the simplest C2D baseline method and promisingly, also lower than all existing state-of-the-arts. 

\newcommand{\specialcell}[2][c]{%
  \begin{tabular}[#1]{@{}c@{}}#2\end{tabular}}
\begin{table}[t]
    \centering
    \caption{Performance evaluated with/without new evaluation protocols (N.E.) and the computation cost of recent methods with DL on MARS~\cite{mars}.}
    \label{tab:neweval}
    \scalebox{0.95}{
    \begin{tabular}{l|c|c|c}
    \hline
    \multirow{2}{*}{Method (\textbf{w/ our DL})} & \multirow{2}{*}{\specialcell{w/o N.E. \\ (mAP)}} & \multirow{2}{*}{\specialcell{\textbf{w/ N.E.}\\ (mAP)}} & \multirow{2}{*}{\#~GFLOPs}  \\ 
                  &   &  &  \\ \hline \hline
    C2D~\cite{ap3d}  & 84.9   & 87.5 & \bf{24.520} \\
    P3D-C~\cite{p3d,ap3d} & 85.0   & 87.5 & 26.030\\
    AP3D~\cite{ap3d}  & 85.4    & 88.2 & 26.369   \\
    TCLNet~\cite{tclnet} & 85.8   & 88.4  &  30.150 \\
    Non-Local~\cite{ap3d,nvan}  & \underline{86.2}  & \underline{88.6}  & 41.733 \\
    \hline
    CF-AAN (\bf{ours})  & \bf{86.5}  & \bf{88.9}& \underline{24.646} \\
    \hline 
    \end{tabular}}
    \vspace{-3mm}
\end{table}

\vspace{-2mm}
\section{Conclusion}
\vspace{-1mm}
In this work, we present a simple re-Detect and Link module to further process the Re-ID datasets,  which can significantly refine the data generated with obsolete methods. Furthermore, the proposed Coarse-to-Fine Axial-Attention Network significantly improves the original non-local module in terms of computational cost with three 1-D position-sensitive axial-attentions and the proposed coarse-to-fine structure while achieving the state-of-the-art performance. 
With our refined data, we find that several baseline models can achieve comparable results with current state-of-the-arts. 
In addition, we also disclose 
the errors not only for the identity labels but also the evaluation protocol for the test data of MARS. With these findings, we hope the release of corrected data 
can encourage the community for the further development of invariant representation on view, pose, illumination, 
and other variations without the hassle of the spatial and temporal alignment and dataset noise.
\vspace{-2mm}
\section*{\normalsize Acknowledgment}
\vspace{-3mm}
\footnotesize This research was supported in part by the Ministry of Science and Technology of Taiwan (MOST 109-2218-E-002 -026), National Taiwan University (NTU-108L104039), Intel Corporation, Delta Electronics and Compal Electronics. 

{\small
\bibliographystyle{ieee_fullname}
\bibliography{egbib}
}

\end{document}